\documentclass[runningheads]{llncs}
\usepackage[T1]{fontenc}
\usepackage{hyperref}
\usepackage{xcolor}
\usepackage{graphicx,verbatim}
\usepackage{array} 

\begin{document}
\title{Post-Processing Methods for Improving Accuracy in MRI Inpainting}
%

\author{Nishad Kulkarni\inst{1}*, Krithika Iyer\inst{1}*, Austin Tapp\inst{1}*,  Abhijeet Parida\inst{1,2},\\ Daniel Capell\'{a}n-Mart\'{i}n\inst{1,2},  Zhifan Jiang\inst{1},  Mar\'{i}a J. Ledesma-Carbayo\inst{2}, \\Syed Muhammad Anwar\inst{1,3},  and Marius George Linguraru\inst{1,3}\\
\email{mlingura@childrensnational.org}}

\authorrunning{N. Kulkarni, K. Iyer, A.Tapp  et al.}
\titlerunning{Post-Processing Methods for Improving Accuracy in MRI Inpainting}
%
\institute{
Sheikh Zayed Institute for Pediatric Surgical Innovation, \\Children’s National Hospital, Washington, DC, USA 
\and 
Universidad Polit\'{e}cnica de Madrid and CIBER-BBN, ISCIII, Madrid, Spain
\and 
School of Medicine and Health Sciences, \\George Washington University, Washington, DC, USA
}

\maketitle              

\begin{abstract}

Magnetic Resonance Imaging (MRI) is the primary imaging modality used in the diagnosis, assessment, and treatment planning for brain pathologies. However, most automated MRI analysis tools, such as segmentation and registration pipelines, are optimized for healthy anatomies and often fail when confronted with large lesions such as tumors. To overcome this, image inpainting techniques aim to locally synthesize healthy brain tissues in tumor regions, enabling the reliable application of general-purpose tools. 
In this work, we systematically evaluate state-of-the-art inpainting models and observe a saturation in their standalone performance. In response, we introduce a methodology combining model ensembling with efficient post-processing strategies such as median filtering, histogram matching, and pixel averaging. Further anatomical refinement is achieved via a lightweight U-Net enhancement stage. Comprehensive evaluation demonstrates that our proposed pipeline improves the anatomical plausibility and visual fidelity of inpainted regions, yielding higher accuracy and more robust outcomes than individual baseline models. 
By combining established models with targeted post-processing, we achieve improved and more accessible inpainting outcomes, supporting broader clinical deployment and sustainable, resource-conscious research. Our 2025 BraTS inpainting docker is available at \href{https://hub.docker.com/layers/aparida12/brats2025/inpt}{hub.docker.com/layers/aparida12/brats2025/inpt}. \\

* These authors contributed equally.

\end{abstract}

\section{Introduction}
Magnetic Resonance Imaging (MRI) and automated analysis are critical for monitoring brain pathologies \cite{bonato2025advancing}. However, existing segmentation and registration tools, designed for healthy anatomies, perform poorly in the presence of large lesions such as glioma, limiting their clinical utility \cite{bonato2025advancing}. To address these limitations, recent efforts have explored the use of image inpainting to locally synthesize healthy-appearing brain tissue in regions affected by tumors. Inpainting, a well-established task in computer vision, involves reconstructing missing or corrupted parts of an image using contextual information \cite{kofler2025brats,bonato2025advancing}. While traditionally applied to 2D natural images, in the context of neuroimaging, inpainting offers the potential to restore anatomical plausibility in 3D MR volumes, enabling the application of standard processing pipelines even in the presence of pathology \cite{zhu2024advancing,santos2025role}. 

The Brain Tumor Segmentation (BraTS) challenge has historically focused on benchmarking algorithms for tumor segmentation in MRI scans of glioma patients \cite{menze2014multimodal,maleki2025analysis}. In 2023, the challenge introduced a new task focused on 3D inpainting of T1-weighted MRIs, where tumor regions were masked out and participants were asked to synthesize realistic, healthy tissue in their place. This task has both clinical and research relevance: in-painted images can improve image registration, facilitate whole-brain parcellation, and support studies of tumor–brain interactions by enabling analysis of the underlying brain anatomy in the absence of visible lesions \cite{santos2025role}. Building on this foundation, the BraTS 2025 inpainting challenge continues to advance the field by encouraging the development of robust, generalizable algorithms for localized 3D brain tissue synthesis. 
 
As the field advances, however, critical questions emerge regarding the direction and efficiency of model development for medical image inpainting. In recent years, deep learning models have reached a plateau, with leading algorithms showing only marginal differences in performance metrics \cite{liu2024lesion,torrado2021inpainting}. Moreover, even with architectural advances, issues like anatomical implausibility, blurred inpainted regions, and imperfect integration with surrounding tissue persist in many models \cite{pollak2025fastsurfer,liu2024lesion}. This diminishing return prompts reflection: does the field benefit from continually training larger, more complex models, or should efforts shift toward more intelligent utilization of existing resources and data? Medical imaging inherently suffers from data scarcity compared to natural image domains, making continued escalation in model complexity impractical and often counterproductive. Overly complex models not only demand significant computational resources but also risk overfitting, reduced interpretability, and high environmental cost factors that collectively limit real-world clinical translation, especially in resource-constrained clinical sites \cite{kamraoui2022longitudinal,guizard2015non}. 

A notable limitation of existing inpainting approaches lies in underutilizing abundant healthy brain scans \cite{liu2020symmetric,iglesias2023synthsr}. Current models often struggle with generating high-fidelity, realistic tissue, occasionally resulting in inpainted images that are blurry or anatomically implausible \cite{sumathi2025high}. In contrast, adopting general purpose yet robust architectures such as denoising autoencoders trained on large, healthy image patches offers a promising, generalizable route for enhancing image quality \cite{he2022masked,sumathi2025high}. These models facilitate the restoration of anatomical consistency without needing large, complex networks by focusing on domain-specific feature extraction from healthy regions. Additionally, traditional image processing techniques provide straightforward, computationally efficient post-processing strategies, including median filtering, pixel averaging (ensemble methods), and sharpening \cite{susan2023deep}. Combined with lightweight learning-based models, these tools have the potential to consistently outperform many computationally intensive approaches, eliminating the need for repeated, large-scale model training \cite{kaur2023complete,dede2018intelligent}. Such solutions could dramatically lower GPU and memory requirements, reduce total energy consumption, and lend themselves to be readily deployable on-site—characteristics that address practical and sustainability considerations for clinical implementation \cite{prajwal2025study}. Additionally, recent studies have demonstrated how intuitively designed methodologies demonstrate robust, generalizable pipelines across diverse clinical environments, as shown in recent efforts leveraging ensemble models for brain tumor segmentation and adaptation in under-resourced settings \cite{parida2024adult,jiang2024enhancing,jiang2024magnetic,capellan2023model}. Thus, streamlined, ensembled methodologies pave the way for building equity in model deployment \cite{ueda2024fairness}. Inequality in access to high-end computational infrastructure is a persistent barrier; therefore, optimizing and reusing established tools democratizes advanced neuroimaging workflows, enabling broader access and more consistent standards of care.

\raggedbottom
\section{Methods}
\begin{figure}
    \centering
    \includegraphics[width=\linewidth]{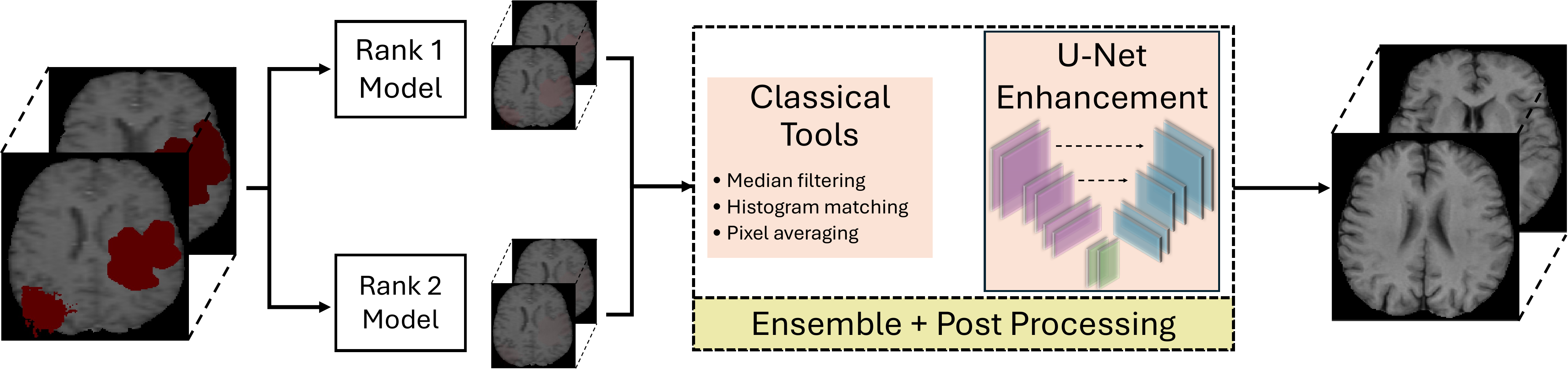}
    \caption{Overview of the proposed post-processing pipeline for MRI inpainting. Inpainted outputs from the top two models from the BraTS 2024 inpainting challenge are first aggregated using classical tools like pixel-level averaging. The ensemble result is then refined by post-processing, a basic U-Net trained on synthetically blurred healthy brain regions, or a mix of both. This multi-stage pipeline improves anatomical fidelity and reconstruction sharpness while maintaining computational efficiency.}
    \label{fig:inpainting_arch}
\end{figure}

\subsection{Dataset}
The BraTS local inpainting dataset (2025) \cite{kofler2023brain} is derived from the BraTS glioma segmentation dataset, which contains multi-modal scans (T1, T1ce, T2, and FLAIR)\cite{menze2015multimodal}.
However, the local inpainting challenge exclusively utilizes the T1-weighted MRI scans for both training and validation. The training set comprises 1251 cases, with each case including the T1 scan, the mask segmenting healthy and unhealthy regions, the full mask, and a voided T1 scan where the lesion region is removed. The validation set consists of 219 cases, each providing only the voided T1 scan and the mask segmenting healthy and unhealthy regions. This structure enables focused development and evaluation of inpainting algorithms on the T1 modality, distinct from segmentation challenges that use all four modalities.

\subsection{Proposed Pipeline}
Our pipeline integrates multiple stages to generate high-quality, healthy brain tissue inpaintings from masked MRI scans, as illustrated in Figure~\ref{fig:inpainting_arch}. Pre-trained models, specifically the top-performing U-Net by Zhang et al.~\cite{zhang2025u} and the 3D Wavelet Diffusion model by Ferreira et al.~\cite{ferreira2024brain}, produce inpainted predictions. These outputs then undergo post-processing steps, including classical pixel averaging filters and a dedicated U-Net-based enhancement module trained to denoise and refine anatomical details. The enhanced outputs from these combined procedures form the final synthesized healthy brain tissue volumes. In the following sections, we describe each component of this pipeline in detail.

\subsubsection{State-of-the-art (SOTA) Inpainting Model Details}
\paragraph{U-Net Based Healthy 3D Brain Tissue Inpainting, J.Zhang et al. \cite{zhang2025u}}
As part of our model ensemble, we incorporate the U-Net-based brain tissue inpainting method developed by Zhang et al. Last year, this model was the top performer in the BraTS Local Synthesis of Tissue via Inpainting challenge, establishing it as the SOTA approach for healthy 3D brain tissue synthesis within masked regions of T1-weighted MRI volumes. This strong benchmark performance makes it a highly robust and justifiable choice for inclusion in our pipeline.

In our approach, we leverage the published implementation and utilize the pre-trained weights, ensuring faithful reproducibility with the original SOTA methodology. The architecture is a sophisticated 3D U-Net with three levels of downsampling and upsampling, skip connections, and a ReLU-activated bridge; regularization techniques include instance normalization and dropout (0.2). The loss function combines mean absolute error within the healthy mask regions and structural similarity index over the entire volume. For our ensemble, masked T1 images are input to the pre-trained U-Net model, with the output inpainted regions seamlessly integrated back into the original image context. 

\paragraph{Conditional Wavelet Diffusion Model, A. Ferreira et al. \cite{ferreira2024brain}}
We incorporate the 3D Wavelet Diffusion Model (WDM) developed by Ferreira et al. as part of our brain tumor inpainting ensemble. This approach achieved second place in the BraTS 2024 Local Synthesis of Tissue via Inpainting challenge, demonstrating competitive performance and complementing the SOTA U-Net model.

The model leverages conditional diffusion processes applied in the wavelet domain, allowing full-resolution 3D brain MRI volumes to be processed without patching or downsampling. This preserves high spatial fidelity during inpainting synthesis while maintaining manageable GPU memory usage. Unlike conventional voxel-space diffusion models, the wavelet transform disentangles spatial frequency components, improving the learning efficiency and reconstruction quality of fine details. The conditional input includes masked MRI images paired with binary masks denoting regions for inpainting. The model was trained to iteratively denoise wavelet coefficients conditioned on these inputs, reconstructing healthy tissue in the tumor-masked regions. 

Inference operates by feeding masked MRI data through the trained wavelet diffusion network to generate consistent, high-fidelity tissue inpaintings and synthesized modalities. The outputs are inverse wavelet transformed back into image space and seamlessly integrated into the original volumes for downstream analysis. By incorporating the pre-trained 3D Wavelet Diffusion Model, our ensemble benefits from an advanced generative approach optimized for high-resolution, volumetric brain MRI synthesis with efficient memory usage and strong inpainting accuracy. Together with the U-Net model, this provides complementary strengths in reproducing realistic healthy brain tissue across diverse pathological scenarios.

\subsubsection{Post-processing} \mbox{}\\[1em]
\noindent\textbf{1. Pixel averaging: median/geometric approach:} After obtaining inpainted outputs from leading SOTA models, classical image processing techniques are employed to enhance visual consistency and reduce instability across the synthesized regions. Specifically, pixel-level aggregation strategies such as voxel-wise average, median, and geometric mean filters are applied to ensemble multiple model predictions. These techniques operate at the voxel level, aggregating pixel intensities across aligned predictions to produce a representative consensus output. Averaging provides a smooth blend of input predictions, minimizing high-frequency disagreement, while the geometric mean is more robust to multiplicative outliers. Notably, when only two predictions are available, average and median aggregations yield mathematically equivalent results. However, the pipeline is designed to accommodate richer ensemble configurations, enabling maximum intensity projection, variance-based adaptive fusion, or hybrid aggregation strategies in future expansions. Collectively, these pixel-level operations offer a robust and interpretable post-processing mechanism that harmonizes multi-model predictions and reduces synthesis variability across diverse MR volumes.

\noindent\textbf{2. Post-averaging smoothing:} We employ two classical denoising strategies to refine the ensembled output. First, we apply a 3D median filter with a \(3 \times 3 \times 3\) kernel, which effectively removes localized outlier voxels while preserving edge structure and anatomical boundaries. This median filter is highly effective when ensembling three or more predictions, such as the top three ranked challenge submissions or hybrid combinations (e.g., two predictions from the first-place model and one from the second-place model). In such cases, the median serves as an outlier-resistant estimator that preserves sharp transitions while rejecting spurious values introduced by any single model and makes it especially beneficial in areas where individual model outputs may hallucinate inconsistent textures or misaligned anatomical boundaries.  Second, we apply Gaussian smoothing with a small standard deviation (\(\sigma = 0.5\)), which gently suppresses high-frequency noise and improves voxel-wise consistency without overly blurring tissue interfaces.

\noindent\textbf{3. Histogram matching:} Following smoothing, we perform histogram matching using the output of the Rank 1 model as the reference. This step aligns the intensity distribution of the ensembled prediction with that of the most reliable single-model output. By correcting for model-specific intensity shifts and ensuring consistency with the original intensity range, histogram matching enhances perceptual realism and improves cross-sample comparability, especially in downstream analyses where intensity drift may affect quantitative accuracy.

\noindent\textbf{4. Deep Learning Enhancement}
As a learned post-processing step, a U-Net model was trained using a synthetic inpainting dataset designed to enable supervised learning of image refinement. To construct this dataset, we started with the original inpainting training dataset and applied random Gaussian blurring to the healthy brain regions defined by the provided healthy masks. This localized degradation mimics the smooth, low-detail appearance of BraTS inpainting methods. The resulting blurred images serve as inputs to the U-Net, while the original, high-resolution MR scans provide the ground truth targets. The network minimizes mean squared error during this task in an attempt to improve ensemble scores. The U-Net effectively minimizes anatomical error, delineates structural boundaries, and optimizes residual smoothing artifacts by leveraging the ground truth healthy tissue appearance. This context-aware enhancement approach complements classical pixel-wise filtering by addressing limitations that ensemble averaging alone cannot resolve. As a result, the U-Net serves as a robust and flexible final-stage enhancement module, significantly improving the perceptual quality and clinical realism of inpainted brain MR volumes.

\section{Experiments}
\subsection{Metrics}
To assess image inpainting quality, we use three main metrics. Structural similarity (SSIM) measures how closely the synthetic image matches the real one in structure, contrast, and luminance; values closer to 1 reflect higher perceptual similarity. Peak signal-to-noise ratio (PSNR) quantifies the ratio between signal and noise based on pixel errors, with higher decibel values indicating better reconstruction quality. Mean squared error (MSE) calculates the average squared difference between predicted and ground truth images, where lower values mean closer pixel-wise agreement. 

We optimized the thresholds and chose the best model with the ranking approach proposed by the BraTS team LaBella \textit {et~al.}~\cite{LaBella2025-oh}. The evaluation is done in a hidden test set, computing lesion-wise metrics in all regions and comparing ranks of all the submissions rather than the metrics. To replicate this procedure, we built an internal ranking metric that produces a single score, where a smaller value means better performance on MSE, SSIM, and PSNR. The code is available on GitHub \footnote{\url{https://github.com/Pediatric-Accelerated-Intelligence-Lab/BraTS-Unofficial-Ranker}}.

The ranking metric approach is robust to outlier predictions, allowing us to optimize for a single value while aligning with the contest evaluation pipeline. 

\subsection{Models for Comparison}
We compare five setups: (1) output from Zhang’s U-Net \cite{zhang2025u}, (2) output from Ferreira’s 3D Wavelet Diffusion \cite{ferreira2024brain}, (3) an ensemble of Zhang’s and Ferreira’s models via geometric or pixel-level averaging, (4) the same ensemble further refined with classical filters such as median or Gaussian, and (5) the filtered ensemble with an additional U-Net-based denoising stage for enhanced image quality. 
Filter implementation followed a straightforward scheme: model outputs from Zhang and Ferreira were combined voxelwise using equal 50/50 weighting, either via geometric averaging (primary setting) or alternative strategies (mean, median, max, min). The refined outputs were subsequently matched to their corresponding references using histogram matching to preserve intensity distributions. 
For the U-Net model, training was conducted for 1000 epochs with 250 iterations per epoch, including 50 validation iterations. We optimized with SGD (momentum = 0.99, Nesterov), weight decay = $3 \times 10^{-5}$, an initial learning rate of $1 \times 10^{-2}$, and polynomial decay. Batch size and patch size were 2 and $96 \times 160 \times 160$, respectively. Simple augmentations of rotations ($\pm 30^\circ$ per axis or anisotropy-aware), uniform scaling [0.7, 1.4], Gaussian noise ($p = 0.1$), Gaussian blur ($\sigma \in [0.5, 1.0]$, $p = 0.2$), brightness scaling ($\times[0.75, 1.25]$, $p = 0.15$), contrast adjustment ($p = 0.15$), simulated low resolution (zoom $\in [0.5, 1.0]$, $p = 0.25$), and gamma corrections ($p = 0.1/0.3$), with mirroring and oversampling of foreground regions were used ($0.33$). The refinement U-Net minimized mean squared error without deep supervision or additional augmentation. Training used a 5-fold cross-validation split, and results were reported as the equal-weight (50/50) ensemble of model probabilities across folds. 
This design enables us to assess the impact of model combination and each post-processing step against the state of the art.

\section{Results}
\begin{figure}[!ht]
    \centering
    \includegraphics[width=\linewidth]{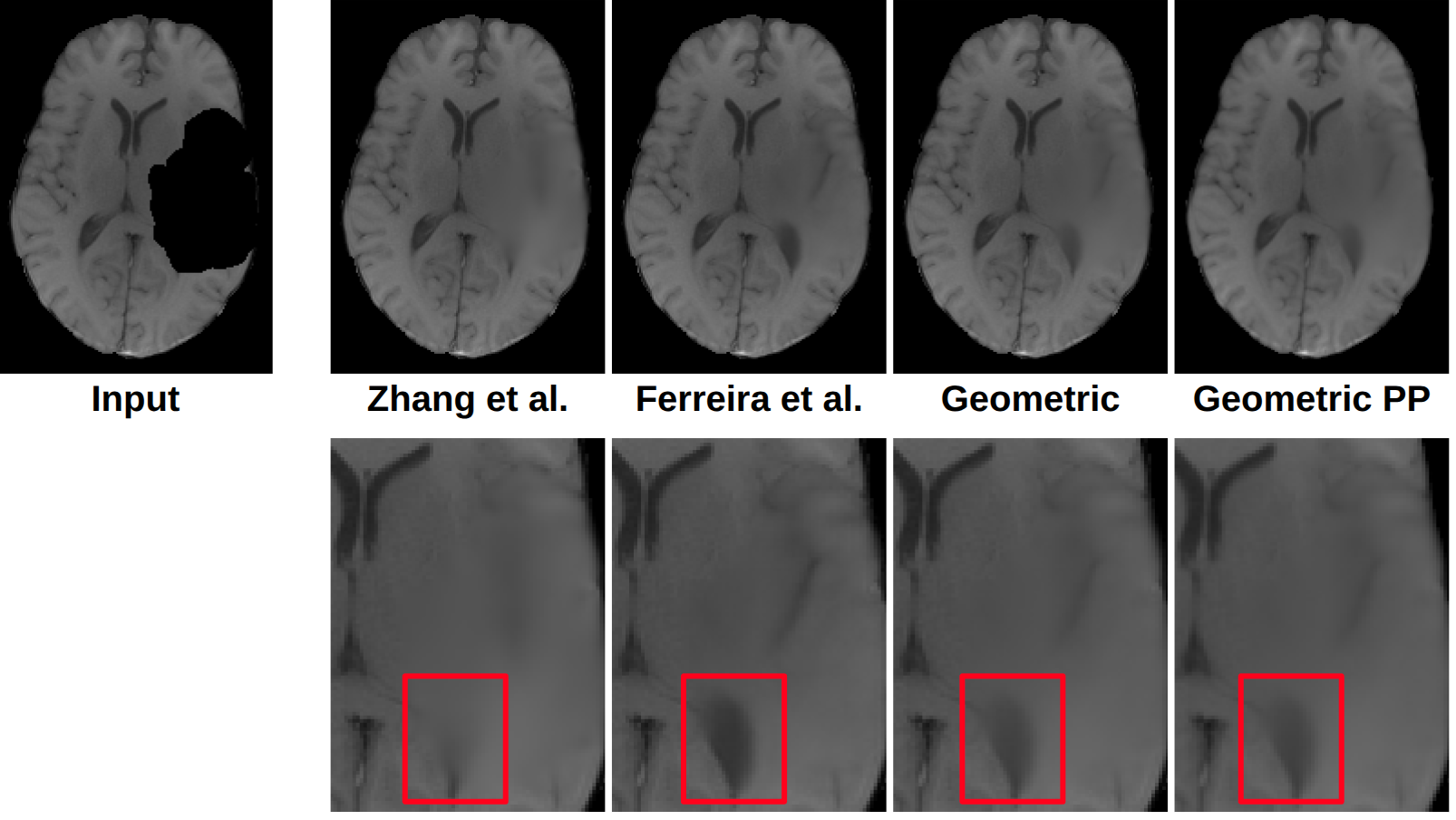}
    \caption{\textbf{Qualitative Results} Comparison of inpainted brain MR images from different methods visualized in ITK-SNAP. From left to right: Rank 1 (Zhang et al.), Rank 2 (Ferreira et al.), median ensemble, and geometric mean ensemble. All views are centered at the same voxel coordinate (164, 110, 78), but demonstrate differing image intensity values under the cursor: 931.1 (Rank 1), 664.2 (Rank 2), 797.6 (Geometric), and 796.4 (Geometric with median filter postprocessing). Notably, the bottom-left ventricle appears more faded and ill-defined in the Rank 1 output, while it is clearly delineated in the Rank 2 image. The ventricle is slightly more apparent in both the median and geometric ensemble outputs than in Rank 1, indicating partial recovery of anatomical detail through pixel-wise fusion strategies. These variations highlight differences in tissue reconstruction fidelity between individual models and ensemble-based approaches.}
    \label{fig:qualitative}
\end{figure}

Table~\ref{tab:val-results-inpainting} presents the quantitative results for our validation-phase experiments using the BraTS 2025 inpainting benchmark. The table summarizes how various post-processing strategies affect the perceptual and quantitative quality of inpainted brain MR volumes. All reported mean~$\pm$~standard deviation values are provided directly by the BraTS validation server, where the standard deviation reflects variation across individual validation cases rather than variation across cross-validation folds. Results on the testing set are included in Table XY.

\begin{table}[!ht]
    \caption{\textbf{Validation Set Quantitative results} on the BraTS 2025 inpainting validation set. Performance is evaluated using Structural Similarity Index (SSIM, $\uparrow$), Peak Signal-to-Noise Ratio (PSNR, $\uparrow$), and Mean Squared Error (MSE, $\downarrow$). Metrics reflect perceptual similarity, reconstruction quality, and pixel-level fidelity, respectively. The best performing configuration for each metric is highlighted in \textbf{bold}.}
    \label{tab:val-results-inpainting}
    \centering
\resizebox{0.95\textwidth}{!}{%
\begin{tabular}{lcccccc}
\hline
\multicolumn{1}{c}{\textbf{Model}} & \textbf{\begin{tabular}[c]{@{}c@{}}Ranking \\ Metric ($\downarrow$)\end{tabular}} & \textbf{MSE($\downarrow$)} & & \textbf{PSNR($\uparrow$)} & & \textbf{SSIM($\uparrow$)} \\ \hline
Rank 1 (Zhang et al.)              & 1.970            & 0.007$\pm$0.005  & &23.257$\pm$4.213  & &0.841$\pm$0.103   \\
Rank 2 (Ferreira et al.)           & 2.441     & 0.007$\pm$0.005  & &22.463$\pm$3.776 & & 0.842$\pm$0.099   \\
Ensemble (R1 + R2)                 & \textbf{1.223}        & \textbf{0.006$\pm$0.004}  & & \textbf{23.652}$\pm$4.092  & &\textbf{0.854$\pm$0.094}   \\
Ensemble + Filters                 & \textbf{1.223}      & \textbf{0.006$\pm$0.004}  & & 23.385$\pm$3.908  & & 0.851$\pm$0.095   \\
Ensemble + U-Net                   & 2.366       & 0.007$\pm$\textbf{0.004 } & &22.740$\pm$\textbf{3.670}    & & 0.843$\pm$0.096   \\ \hline
\end{tabular}
}
\end{table}

\begin{table}[!ht]
    \caption{\textbf{Test Set Quantitative results} on the BraTS 2025 inpainting testing set. Performance is evaluated using Structural Similarity Index (SSIM, $\uparrow$), Peak Signal-to-Noise Ratio (PSNR, $\uparrow$), and Mean Squared Error (MSE, $\downarrow$). Metrics reflect perceptual similarity, reconstruction quality, and pixel-level fidelity, respectively. }
    \label{tab:test-results-inpainting}
    \centering
\resizebox{0.80\textwidth}{!}{%
\begin{tabular}{>{\raggedright\arraybackslash}p{0.3\linewidth}llccccc}
\hline
\textbf{Model} &   &&\textbf{MSE($\downarrow$)} & & \textbf{PSNR($\uparrow$)} & & \textbf{SSIM($\uparrow$)} \\ \hline
Ensemble + Filters                   &                               &&0.007$\pm$0.004 & &23.955$\pm$4.989& & 0.867$\pm$0.1312\\ \hline
\end{tabular}
}
\end{table}

The baseline Rank 1 and Rank 2 models (Ferreira et al. \cite{ferreira2024brain} and Zhang et al. \cite{zhang2025u}, respectively) yield identical values for MSE (0.007$\pm$0.005) and nearly identical SSIM (0.841$\pm$0.103 vs. 0.842$\pm$0.099), but differ in PSNR: 23.257$\pm$4.213 for Rank 1 and 22.463$\pm$3.776 for Rank 2. Simple ensembling of these models yields strong improvements across all metrics, with MSE reduced to 0.006$\pm$0.004, PSNR increased to 23.652$\pm$4.092, and SSIM elevated to 0.854$\pm$0.094. Importantly, this approach also lowers the standard deviations for all metrics, suggesting more stable and consistent performance. Adding classical post-processing filters preserves the MSE and maintains low variance, but leads to slightly reduced PSNR (23.385$\pm$3.908) and SSIM (0.851$\pm$0.095) relative to ensembling alone.

In contrast, the U-Net-enhanced ensemble results in performance that, while still stable, shows a decline in mean values: MSE increases to 0.007$\pm$0.004, PSNR drops to 22.740$\pm$3.670, and SSIM falls to 0.843$\pm$0.096. According to the overall ranking metric, the best configuration is the ensemble with classical filters (1.223), followed by Rank 1 (1.970), then the U-Net-enhanced ensemble (2.366). These results suggest that while the U-Net refiner introduces plausible structural detail and reduced variability, it may require further training to outperform simpler, well-tuned classical strategies.
\section{Discussion}\raggedbottom

Motivated by the observed saturation in performance among state-of-the-art inpainting models, we proposed a modular pipeline that combines predictions from previous winners' methods with classical image processing filters and a U-Net-based enhancement module trained on synthetically degraded healthy tissue. The synthetically degraded tissue uses a simple method that applies a Gaussian blur to healthy brain regions, and using these artificially degraded scans as inputs, we generated a rich supervised training dataset for the enhancement model without additional manual annotations. This allowed the U-Net to learn a targeted correction function aimed at restoring anatomical similarity in artifact-prone outputs from existing inpainting models. 

Our results indicate that simple model ensembling leads to clear performance improvements across SSIM, PSNR, and MSE while reducing variability compared to individual model outputs. Adding classical post-processing filters preserves these gains with minimal overhead, and ranks best overall across quantitative metrics. Our metrics reflect a third place ranking in the BraTS 2025 inpainting challenge with a strategy that emphasizes using an ensemble of existing models and lightweight post-processing rather than creating and training increasingly complex architectures. For the 2025 validation set, the 2024 baseline Rank-1 (Ferreira et al. \cite{ferreira2024brain}) and Rank-2 (Zhang et al. \cite{zhang2025u}) systems yield identical MSE ($0.007\pm0.005$) and near-identical SSIM ($0.841\pm0.103$ vs. $0.842\pm0.099$), but differ in PSNR ($23.257\pm4.213$ vs. $22.463\pm3.776$). A simple ensemble of these models produces consistent improvements across all metrics (MSE $0.006\pm0.004$, PSNR $23.652\pm4.092$, SSIM $0.854\pm0.094$) and meaningfully reduces standard deviations, indicating more stable performance across slices/cases. Adding classical filters preserves the ensemble's MSE and low variance but slightly lowers means (PSNR $23.385\pm3.908$, SSIM $0.851\pm0.095$). The U-Net-based refinement showed slight reductions in mean SSIM and PSNR, along with an increase in MSE. These results suggest that while conceptually promising and effective at maintaining low standard deviation, the U-Net enhancement model in its current training configuration may require additional fine-tuning or regularization to outperform traditional averaging and filtering approaches. 

Importantly, the components of our postprocessing pipeline are computationally lightweight, reproducible, and deployable in resource-constrained settings. This efficiency not only reduces the burden on GPU memory and energy consumption but also makes high-fidelity inpainting more accessible across diverse research and clinical environments. By strategically leveraging existing models, healthy data cohorts, and simple yet effective post-processing methods, our approach offers a sustainable and modular path forward for medical image inpainting and supports bridging the gap between benchmark performance and real-world applicability.

\section{Conclusion}
Our work shows that strong brain tissue inpainting performance can be achieved without resorting to ever larger or more complex models. By ensembling established, high-performing models and applying lightweight classical post-processing tools alongside a U-Net-based enhancement module trained on healthy scans, we obtain high-fidelity results in a resource-efficient manner. These approaches do not require extensive GPU resources and are simple to implement, demonstrating that carefully selected and combined methods, rather than sheer model scale, can provide reliable, clinically useful brain image synthesis. This strategy supports scalable and accessible solutions for medical image inpainting in both research and clinical practice.

%
%
\bibliographystyle{splncs04}
\bibliography{ref}

\end{document}